\documentclass[runningheads]{llncs}

\usepackage{graphicx}
\usepackage{amssymb}
\usepackage{rotating}

\usepackage{float}
\usepackage{subfigure}

\usepackage{subfigure}
\usepackage{wrapfig}
\usepackage{booktabs}
\usepackage{multirow}
\usepackage{makecell}
\usepackage{enumitem}

\usepackage{bbding}
\usepackage{xcolor}

\usepackage{bm}
\usepackage{amsmath}
\usepackage{amssymb}
\usepackage{mathtools}
\usepackage{xspace}
\usepackage{multirow}

\usepackage[capitalise]{cleveref}

\begin{document}

\title{GANExplainer: GAN-based Graph Neural Networks Explainer}
\titlerunning{GANExplainer}

\author{Yiqiao Li \and Jianlong Zhou \and Boyuan Zheng \and Fang Chen}

\institute{University of Technology Sydney, Sydney, Australia\\
\email{yiqiao.li-1@student.uts.edu.au}\\
\email{jianlong.zhou@uts.edu.au}\\
\email{boyuan.zheng-1@student.uts.edu.au}\\
\email{Fang.Chen@uts.edu.au}
}

\maketitle             
\begin{abstract}
With the rapid deployment of graph neural networks (GNNs) based techniques into a wide range of applications such as link prediction, node classification, and graph classification the explainability of GNNs has become an indispensable component for predictive and trustworthy decision-making. Thus, it is critical to explain why graph neural network (GNN) makes particular predictions for them to be believed in many applications. Some GNNs explainers have been proposed recently. However, they lack to generate accurate and real explanations. To mitigate these limitations, we propose GANExplainer, based on Generative Adversarial Network (GAN) architecture. GANExplainer is composed of a generator to create explanations and a discriminator to assist with the Generator development. We investigate the explanation accuracy of our models by comparing the performance of GANExplainer with other state-of-the-art methods. Our empirical results on synthetic datasets indicate that GANExplainer improves explanation accuracy by up to 35\% compared to its alternatives.
\end{abstract}

\section{Introduction}\label{sec:intro}
Graph neural networks (GNNs)~\cite{wu2020comprehensive}, with the resurgence of deep learning, have become a powerful tool to model graph datasets and achieved impressive performance. However, a GNN model is typically very complicated and how it makes predictions are unclear; while unboxing the working mechanism of a GNN model is crucial in many practical applications (e.g., criminal associations predicting \cite{wang2022hagen}, traffic forecasting \cite{jiang2022graph}, and medical diagnosis \cite{chen2021causal}). Thus, an explainable model is favoured and even necessary, as explanations benefit users in multiple ways, such as improving the model's fairness/security and it also enhances understanding and trust in the model's predictions. As a result, explaining GNNs has achieved considerable research attention in recent years.

Recently, several explainers~\cite{abs-1903-03894,huang_graphlime:_2022,yuan_xgnn:_2020,LuoCXYZC020,lin2021generative} have been proposed to tackle the problem of explaining GNN models. These attempts can be categorized into \emph{local} and \emph{global} explainers according to their interpretation scales. In particular, if the method only provides an explanation for a specific instance, it is a \emph{local explainer}. In contrast, if the method explains the whole model, then it is a \emph{global explainer}. Alternatively, GNN explainers can also be classified as either \emph{transductive} or \emph{inductive} explainers based on their capacity to generalize to extra unexplained nodes. GNNExplainer~\cite{abs-1903-03894} and GraphLIME~\cite{huang_graphlime:_2022} are challenging to be applied in inductive settings as their explanations are limited to a single instance, and they provide local explanations, which are incapable of capturing the archetypal patterns shared by the same classes or groupings. While PGExplainer~\cite{LuoCXYZC020}, Gem~\cite{lin2021generative}, and XGNN~\cite{yuan_xgnn:_2020} can provide a global explanation of the model prediction. Specifically, a trained PGExplainer~\cite{LuoCXYZC020} and Gem~\cite{lin2021generative} can be used in inductive scenarios to infer explanations for unexplained instances without retraining the explanation models. 

However, XGNN trains a graph generator for explaining a class by outputting class-wise graph patterns. It hardly applies to a specific instance, since the graph patterns may not even exist on the instance. PGExplainer trained a shared generative probabilistic model using the multiple explained instances rather than explicitly dissecting and modelling the class-wise knowledge. Gem trained an auto-encode generative model based on Granger causality. Still, the graph dataset's explanation is inconsistent since it lacks a discriminator to monitor the generation progress.

To address the above limitations, we propose a global inductive explainer, Generative Adversarial Explainers (GANExplainer) using the virtues of Generative Adversarial Network (GAN)~\cite{GAN_2014}. GAN application widely arranged from image generation to 3D object generation, but not used in GNN explaining. We are the first methods to use GAN to generate explanations for GNNs. Inspired by the explanations generated by other explainers, our ultimate goal is to encourage a compact subgraph of the computation graph to have a large causal influence on the outcome of the target GNN. Our setting is general and works for any graph learning tasks, including node classification and graph classification. Our contributions can be summarized as followings:
\begin{itemize}
    \item We endeavour to apply GAN to a new domain. GAN has been implemented in numerous fields, including computer vision~\cite{VondrickPT16}, image processing~\cite{RadfordMC15}, and natural language processing~\cite{QiaoZXT19}. However, no attempts are made to explain GNNs using GAN. We are the first to use a GAN to shed light on how GNNs process and learn from graph data, thereby enhancing our understanding of their inner workings.
    
    \item  We present an innovative GNNs Explainer. Inspired by the GAN architecture, we propose a GANExplainer that is supervised by the target GNN and can consistently provide accurate explanations. GANExplainer is composed of both a Generator and a Discriminator. The objective of the Generator is to generate explanations that can be fed into the target GNN to obtain the same predictions. The Discriminator is a graph classifier that category the generating and input graphs into distinct categories.
    
    \item We enhance the accuracy of GNNs Explainers.  GANExplainer is not only capable of generating global explanations, but can also be utilised in inductive settings. Compared to state-of-the-art inductive GNN explainers, GANExplainer has superior performance.
    
\end{itemize}

\section{Related Work}
\paragraph{Generate Adversarial Network.} Generative adversarial networks (GANs)~\cite{GAN_2014} are a type of deep learning model that can generate new data samples similar to a training dataset. They consist of two neural networks: a generator and a discriminator. The generator tries to capture the distribution of actual examples and generate new data examples. The discriminator is usually a binary classifier used to discriminate generated examples from actual examples as accurately as possible. The optimization of GANs is a minimax optimization problem. The optimization terminates at a saddle point, forming a minimum for the generator and a maximum for the discriminator. That is, the GAN optimization goal is to reach Nash equilibrium. At that point, the generator can be considered to have accurately captured the distribution of real examples. 

GANs provide a way to learn deep representations without extensively annotated training data. They achieve this by deriving backpropagation signals through a competitive process involving a pair of networks. The representations that can be learned by GANs may be used in a variety of applications, including video generation~\cite{VondrickPT16}, image generation~\cite{YangKBP17}, face generation~\cite{GAN_2014}, object detection~\cite{EhsaniMF18}, and texture synthesis~\cite{LiW16}. However, utilizing GAN in explaining GNNs is still under exploration.

\paragraph{GNNs Explainers.} GNNs incorporate both graph structure and feature information, which results in complex non-linear models, rendering explaining its prediction remain a challenging task. Besides, model explanations could bring a lot of benefits to users (e.g., improving safety and promoting fairness). Thus, some popular works have emerged in recent years focusing on the explanation of GNN models by leveraging the properties of graph features and structures. We here briefly review the respective GNNs explainers below.

\emph{GNNExplainer}~\cite{abs-1903-03894} is a seminal method in the field of explaining GNN models. It provides local explanations for GNNs by identifying the most relevant features and subgraphs, which are essential in predicting a GNN.~\emph{PGExplainer}~\cite{LuoCXYZC020} introduces explanations for GNNs with the use of a probabilistic graph. It provides model-level explanations for each instance and possesses strong generalizability. CF-GNNExplainer~\cite{DBLP_LucicHTRS22} generate counterfactual explanations for the majority of instances for GNN explanations but ignores the correlation between the prediction and the explanation. Thus, Bajaj et al.~\cite{BajajCXPWLZ21} proposed RCExplainer generating robust counterfactual explanations. And Wang et al.~\cite{multi_grained_WangWZHC21} proposed ReFine, which pursues multi-grained explainability by pre-training and fine-tuning. 

Also, reinforcement learning is prevalent in explaining GNNs. Such as, Yuan et al.~\cite{XGNN_YuanTHJ20} proposed XGNN, which is a model-level explainer that trains a graph generator to generate graph patterns to maximize a specific prediction of the model. Since XGNN focuses on model-level explanations, it may not preserve the local fidelity, which means its explanations may not be a substructure existing in the input graph. Thus, to solve this limitation, Wang et al.~\cite {RC_Explainer_Wang} proposed RC-Explainer, which generates causal explanations for GNNs by combining the causal screening process as a Markov Decision Process in reinforcement learning. Further, Shan et al.~\cite{RGExplainer_ShanSZLL21} proposed RG-Explainer, a reinforcement learning enhanced explainer that can be applied in the inductive setting, demonstrating its better generalization ability. 

~\emph{Gem}~\cite{lin2021generative} is able to provide both local and global explanations, and it is also operated in an inductive setting. Thus, it can explain GNN models without retraining. Notably, it adopts a parameterized graph auto-encoder with Graph Convolutional Network(GCN)~\cite{kipf2016semi} layers to generate explanations. Also, Gem applies Granger causality to generate causal explanations. However, Gem is not consistently getting accurate explanations. Thus, we aim to improve it by adding a discriminator in our framework, which can provide high-accuracy explanations in synthetic and real-world datasets. 
\section{Methods}
\subsection{Problem Formulation}
GNN explainer provides a faithful and compact subgraph to illustrate why the GNNs make the prediction, which indicates essential graph structures and features leading to the model outcomes. Alternatively, it is important to note that the explanation subgraph must be a real subgraph of the input graph, meaning it must contain a subset of the vertices and edges of the input graph.
 \begin{figure*}[!htpb]
    \centering
    \includegraphics[width=0.5\textwidth]{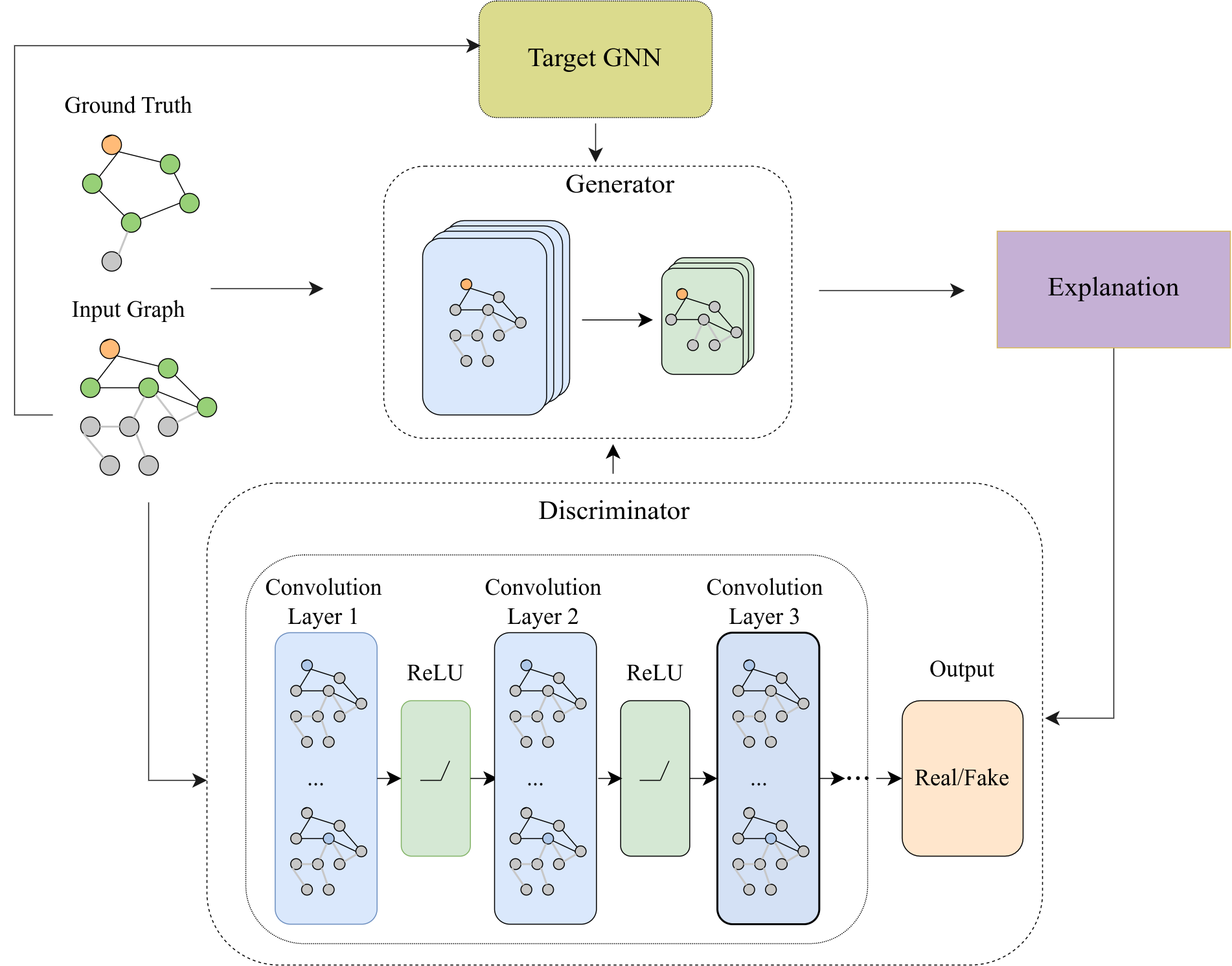}
    \caption{The framework of GANExplainer. We generate ground truth through the Gem distillation process.}
    \label{fig:framework}
\end{figure*}

For a graph $\bm{g}=(\bm{V},\bm{A},\bm{X})$ with label $\bm{L}=\{l_1,l_2,...,l_i\}$, $l \in C=\{c1, c2, ...,c_j\}$ and $j$ is the number of categories. Where $\bm{A}$ presents the adjacency matrix, when node $i$ and node $j$ are connected, $A_{ij}=1$, if node $i$ and node $j$ are not connected, then $A_{ij}=0$. We have the prediction $f(\bm{g})=y$ of a GNN model, and the explanation $E(f(\bm{g}), \bm{g})=Exp$ from a GNN explainer. We expect putting the explanation from the GNNs explainer $Exp$  to the target model $f(\bm{g})$ can obtain the same prediction. Shortly, we expect obtaining $f(\bm{g})=y$ and $f(E(f(\bm{g}), \bm{g})) = y$. Alternatively, we expect the explanation to be the subgraph of the input graph, that is $Exp \in \bm{g}$.

Thus, to provide an accurate and real explanation, We propose a GANExplainer based on GAN architecture, which thoroughly applies the feature and structure of the graph together. Applying the feature and structure of a graph to the GAN architecture could potentially allow the GANExplainer to more accurately capture the underlying relationships and patterns in the data, which could lead to more accurate explanations. 

\subsection{GANExplainer}
Inspired by the GAN architecture, we propose the GANExplainer, which attempts to generate explanations for GNN under the supervision of the target GNN or ground truth. The framework of GANExplainer is shown in Figure~\ref{fig:framework}. 

To keep the explanation generated by GANExplainer as a substructure of the input graph, we assign weights to each edge to determine its relative importance. Suppose the explanation adjacency matrix $\bm{Exp}$ contains elements that are not present in the adjacency matrix $\bm{A}$ of the input graph. In that case, the explanation subgraph is not a real subgraph of the input graph. Our objective is to generate real explanations that only contain truth edges of the input graph. Thus, we use GANExplainer to produce a weight matrix $\bm{W}$ instead of generating a subgraph directly. When we get a weight matrix $\bm{W}$, we multi the weight matrix into adjacency matrix $\bm{A}$, and we get the explanation adjacency matrix $\bm{Exp}$ ($\bm{Exp}=\bm{W}*\bm{A}$).  $\bm{Exp}$ represents the explanation subgraph. The element at the $i-th$ row and $j-th$ column of the explanation adjacency matrix is the weight of the edge between the $i-th$ and $j-th$ vertices in the explanation subgraph.

\paragraph{Generator.} The purpose of the generator is to create a weighted subgraph. The Generator is composed of an encode and a decord section. Particularly, for synthetic datasets, the Generator employs a 6-layer encode and a 2-layer decode, while for real-world datasets, it employs a 7-layer encode and a 2-layer decode. The Generator initially produces a weighted graph based on the input graph, the ground truth of the input graph, and the prediction of the target GNN.  The product is then introduced into the Discriminator. Then, based on the feedback from the Discriminator, the Generator was modified. 

\paragraph{Discriminator.} The objective of the Discriminator is to identify whether the graph is real or artificial. The discriminator is a three-layer graph classifier with convolutional layers. In specific, the output of the Discriminator for the input graphs is real labels. In comparison, the discriminator produces fake labels for the graphs generated by the Generator, which information will be fed back into Generator and assist with the Generator development.

\subsection{Improved Loss Function}
GANs consist of a generator and a discriminator. The generator and the discriminator interact to get a balance like in a minimax game. The objective of the generator $G$ is to generate data that tricks the discriminator, whereas the objective of the discriminator $D$ is to differentiate between actual and generated data. Consequently, the loss function of the original GAN is defined as follows:

\begin{align*}
\min _G \max _D V(D, G)=E_{x \sim p_{\text {data }}(x)}[\log D(x)]\\
+E_{z \sim p_z(z)}[\log (1-D(G(z)))]
\end{align*}

where the $x$ is the input graph data, and the $z$ is the normalized graph adjacency matrix.

Explainers should provide explanations for GNNs by finding both significant subgraphs and essential features that play a crucial role in the prediction of GNN. Therefore, we must use the target GNN as a check to ensure that we generate explanations that can explain the GNN, and that these explanations are not limited to the critical structures of datasets. Consequently, if we only use the original GAN objective function, we will generate another graph/subgraph that cannot explain why the GNN makes the prediction. In order to generate explanations with two virtues fidelity and reality, we have enhanced the generator's objective function. The objective function of the Generator of GANExplainer is defined as follows:
\begin{align*} 
\min _{G} {_{z \sim p_{z(z)}}}[{\log (1-D(G(z))) } + \lambda {\frac{1}{N} \sum_{i=1}^N\left(f(\bm{g})-f(G(z))\right)^2}]
\end{align*}
\noindent
where the $f$ means the target GNN; the $N$ is the node set of $\bm{g}$. In our experiments for synthetic datasets, we set $\lambda=2$, while for real-world datasets we set $\lambda=3$.

\section{Experiments}
In this section, we conduct experiments to inspect the performances of our model. We first describe the details of the datasets and implementation we used in~\ref{subsec:dataset} and ~\ref{subsec:imdetails} respectively. After that, we present and analyze the experimental results on synthetic datasets in~\ref{subsec:ressyn} and real-world datasets in~\ref{subsec:reseal} from two aspects: quantitative evaluation and qualitative evaluation.

\begin{table*}
\caption{Datasets information.}
\centering
\setlength{\tabcolsep}{4.0mm}{
\begin{tabular}{l c c  c c}
\toprule
& \multicolumn{2}{c}{Node Classification}& Graph Classification\\
\cmidrule(l){2-3}
\cmidrule(l){4-5}
& BA-Shapes & Tree-Cycles & Mutagenicity & NCI1\\
\midrule
\# of Graphs & 1 & 1  & 4,337 & 4110\\
\# of Edges & 4110 & 1950   & 266,894 & 132,753\\
\# of Nodes & 700  & 871   & 131,488 & 122,747\\
\# of Labels & 4 & 2  & 2 & 2\\    
\midrule
\end{tabular}
}
\label{tab:dataset}
\end{table*}
\subsection{Datasets}\label{subsec:dataset}
We focus on two widely used synthetic node classification datasets, including BA-Shapes and Tree-Cycles~\cite{abs-1903-03894,GNNs_Explainers_Robustness}, and two Real-world graph classification datasets, Mutagenicity~\cite{kazius2005derivation} and NCI1~\cite{WaleWK08}. Statistics of these datasets are shown in Table~\ref{tab:dataset}. 

The BA-Shapes dataset is based on a Barabasi-Albert (BA) graph on 300 nodes and attaches 80 "house"-structured network motifs to randomly selected nodes of the base graph. Nodes are categorised into four classes depending on their structural roles, which correspond to nodes at the top, middle, and bottom of houses and nodes that do not belong to a house. 

The Tree-Cycles dataset starts with a base 8-level balanced binary tree and attaches 80 six-node cycle motifs, which are attached to random nodes of the base graph. Nodes are divided into 2 classes according to nodes belong the tree or the cycle. 

Mutagenicity datasets contain 4337 molecule graphs, where nodes represent atoms, and edges denote chemical bonds. The graph classes, including the non-mutagenic or the mutagenic class, indicate their mutagenic effects on the Gram-negative bacterium Salmonella Typhimurium. Carbon rings with chemical groups $NH_2$ or $NO_2$ are known to be mutagenic. Carbon rings, however, exist in both mutagen and nonmutagenic graphs, which are not discriminative.

NCI1 represents a balanced subset of chemical compounds screened for activity against non-small cell lung cancer. This dataset contains more than 4,000 chemical compounds, each of which has a class label between positive and negative. Each chemical compound is represented as an undirected graph where nodes, edges and node labels correspond to atoms, chemical bonds, and atom types respectively.

\subsection{Experimental Settings}\label{subsec:imdetails}
In our experiments, we replicate the experimental conditions of Gem so that we can fairly compare our results to those of Gem, which are our baseline. In accordance with experimental Gem settings, they select K to select various top edges as explanations. Consequently, we set K to the same value for each dataset as the Gem. Consistent with experimental Gem settings, we divide the data into 80\% training data, 10\% validation data, and 10\% testing data. We maintain the consistency of the testing data. Both training data and validation data are utilised in their entirety during the training process.

\paragraph{Baseline approaches.} With the wide application of GNNs, more and more GNN explainers have been proposed to address the problem of explaining GNN models. \emph{GNNExplainer} is a seminal method in the field of explaining GNN models. In addition, the PGExplainer and Gem are most related to our method. Note that for the synthetic datasets, we know the motif of each dataset. However, for real-world datasets, there are no explicit motifs (no ground truth motifs) for classification. Thus, for real-world datasets, we need to explain all classes. Since PGExplainer assumes NO2 or NH2 as the motifs for the mutagen graphs and trains an MLP for model explanation with the mutagen graphs including at least one of these two motifs. In addition, as the results reported by the Gem, we note that the performance of Gem is better than PGExplainer. Thus, we consider GNNExplainer and Gem as alternative approaches. We set all the hyperparameters of the baselines as reported in the corresponding papers.

\paragraph{Metrics.}
A better explainer should be able to generate more compact subgraphs yet maintains the prediction accuracy while the associated explanations are fed into the target GNN. After comparing the characteristic of each metric~\cite{yiqiaosurvey,zhou2021evaluating}, we chose quantitative and qualitative evaluation. To this end, we generate the explanations for the test set based on GNNExplainer, Gem, and GANExplainer, respectively. Then we use the predictions of the target GNN for the explanations to calculate the explanation accuracy. The explanation accuracy can be defined as:
\begin{align*} 
ACC_{exp} = \frac{Correct_{f(\bm{g})=f(Exp)}}{|Test|}
\end{align*}

where the $f$ means the target GNN; $\bm{g}$ presents the graph; $Exp$ presents the explanation; $|Test|$ means the total number of test set.

\begin{table*}[]
\caption{Results on Synthetic Datasets.}
\centering
\begin{tabular}{llllll|lllll}
\midrule
\multicolumn{1}{c}{K (edges)} & \multicolumn{5}{|c|}{BA-Shapes} & \multicolumn{5}{c}{Tree-Cycles} \\
\multicolumn{1}{c|}{} & 5    & 6    & 7    & 8   & 9   & 6    & 7    & 8    & 9   & 10   \\ 
\midrule
GNNExplainer &   0.7941
  &   0.8824
  & 0.9118
 & 0.9118
&  0.9118
  & 0.2000 &0.5429
 & 0.7143
 &  0.8571
 &  0.9429  \\
Gem   & 0.9412& 0.9412
& 0.9412
 & 0.9412
 &0.9412
& 0.7429
&0.7429
 & 0.7714
 & 0.8857
 &0.9143
 \\
GANExplainer   & 0.7647
 &1.000 &0.9706
 &0.9853
 &  0.9853&  0.9143
   &  1.0000  &    0.9714
  &   1.0000
  & 1.0000
   \\
\midrule
\end{tabular}
\label{tab:res_synthetic}
\end{table*}

\subsection{GANExplainer on Synthetic Datasets}\label{subsec:ressyn}
Firstly, we conduct experiments on synthetic datasets, including BA-Shapes and Tree-Cycles. We evaluate the accuracy of explanations provided by GANExplainer (our model), Gem, and GNNExplainer. We also present quantitative and qualitative evaluations of our experiments.

The accuracy of explanations for synthetic datasets with various K settings is detailed in Table~\ref{tab:res_synthetic}. The results indicate that GANExplainer consistently provides the most accurate explanations in all cases. On the BA-Shapes dataset, GNNExplainer, Gem, and GANExplainer perform well for synthetic datasets. However, GANExplainer also incorporates a number of enhancements. GANExplainer outperforms GNNExplainer and Gem on BA-Shapes. On the Tree-Cycles dataset, GANExplainer performs well on Tree-Cycles, whereas neither GNNExplainer nor Gem perform well. Specifically, when K=7 on Tree-Cycles, GANExplainer achieves a 35\% and 84\% improvement over Gem and GNNExplainer, respectively. 

\begin{figure*}[!htpb]
    \centering
    \includegraphics[width=1.0\textwidth]{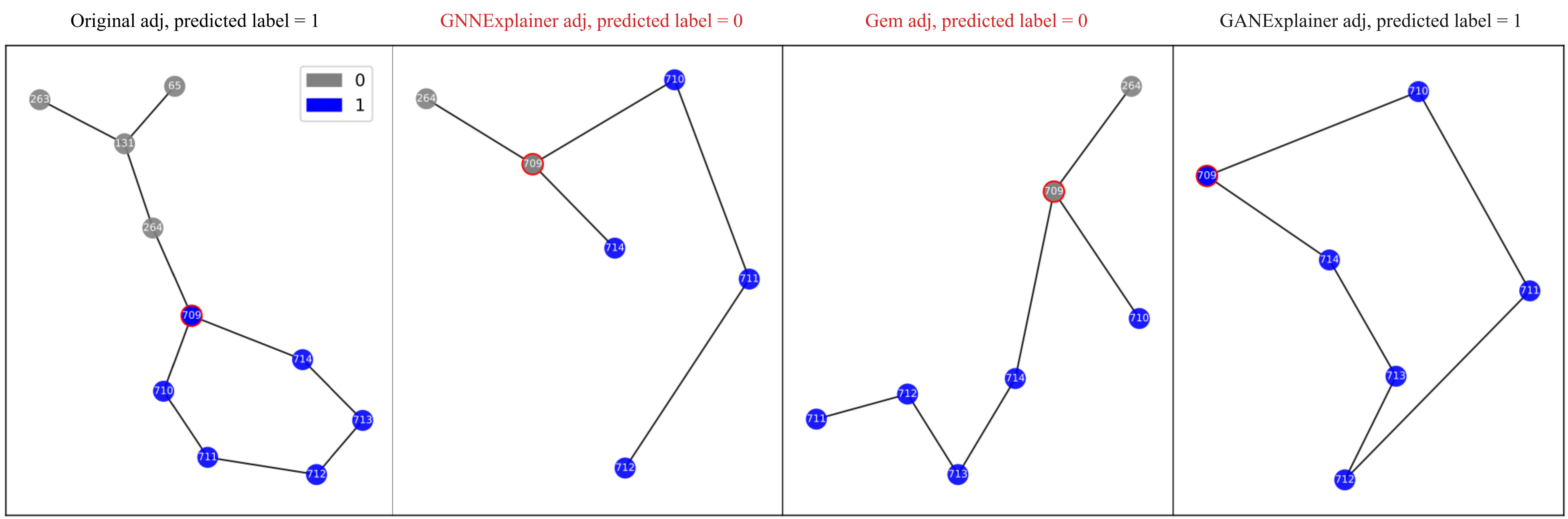}
    \caption{The explanation visualization of Node 709, the label is 1, on Tree-Cycles when $K=6$. In this diagram, the blue nodes mean the prediction label of the target GNN is 1, and the grey nodes mean the predictions of the target GNN for these nodes are 0. The red circle node is the node that needs to explain why the target GNN predicting label is 1. The first subfigure is the original graph structure, and the prediction of the target GNN is 1, which means the target GNN makes the right prediction. The second to fourth subfigures are the explanation of the node 709 from GNNExplainer, Gem, and GANExplainer, respectively.}
    \label{fig:res_syn4}
\end{figure*}

Qualitative is an effective way to visualize the explanations. We obtain the difference in explanations between GNNExplainer, Gem and GANExplainer by qualitative analysis. We visualise the explanations of Tree-Cycles when $K=6$, shown in Figure~\ref{fig:res_syn4}. We note that the explanations of GNNExplainer and Gem can not get the correct prediction when fed into the target GNN. However, we can get the correct prediction similar to the prediction of the target GNN when fed the explanation from our explainer, GANExplainer.
\begin{table*}[]
\caption{Results on Real-world Datasets.}
\centering
\begin{tabular}{lllll|llll}
\midrule
K (egeds) & \multicolumn{4}{|c|}{Mutagenicity} & \multicolumn{4}{c}{NCI1} \\
\multicolumn{1}{c|}{} & 15    & 20   & 25    & 30   & 15  & 20   & 25   & 30\\ 
\midrule
GNNExplainer &0.6981& 0.7188& 0.7442 &0.7834 &0.6909&0.7031 &0.7566&0.8004 \\
Gem  & 0.6705& 0.7027& 0.7741 & 0.7949 &0.6253&0.7055&  0.7956& 0.8126  \\
GANExplainer  & 0.6935  &  0.7442 & 0.7650 &0.7857 &0.6642 &  0.7494&0.7908 & 0.8273 \\
\midrule
\end{tabular}
\label{tab:res_realworld}
\end{table*}

\subsection{GANExplainer on Real-world Datasets}\label{subsec:reseal}
We report the results of real-world datasets in the following. The quantitative evaluation is shown in Table~\ref{tab:res_realworld}. As shown in the table, We note that the results reported successfully verify that our proposed GANExplainer can generate explanations that consistently yield high explanation accuracies over all datasets. 

\begin{figure*}[!htpb]
    \centering
    \includegraphics[width=1.0\textwidth]{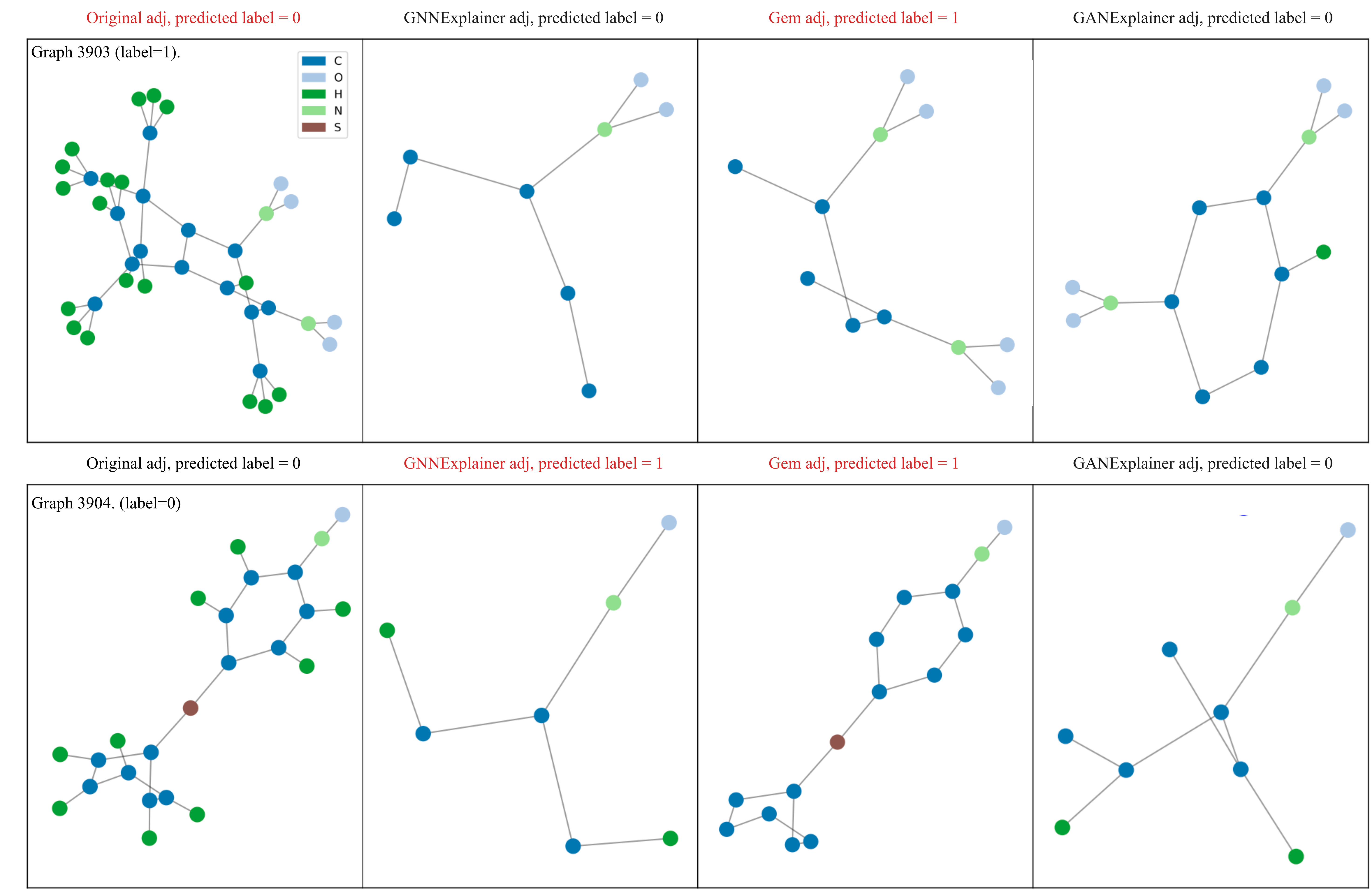}
    \caption{The explanation visualization on Mutagenicity when $K=15$. Graph 3903 the label is 1, and the graph 3904 the label is 0. In this diagram, the first column is the original graph structure and the predicted label of the target GNN. The second to fourth columns are the explanation of graphs from GNNExplainer, Gem, and GANExplainer, respectively.}
    \label{fig:res_mutag}
\end{figure*}

To further check the explainability of the generated explanations, we report the qualitative evaluation of the Mutagenicity (graph 3903 and graph 3904, $K=15$) in Figure~\ref{fig:res_mutag}. 

When the target GNN gets the prediction of the graph, we expect to get the same prediction of explanation. Specifically, for graph 3903, the label is 1, while we get the wrong prediction of the target GNN 0. We want to explain why the target GNN makes the prediction 0. Thus, we visualise the explanations from GNNExplainer, Gem and GANExplainer, respectively. From the figure, we note that the explanations of GNNExplainer and GANExplainer get the same prediction as the original graph, after feeding explanations into the target GNN. However, the explanation of Gem makes a different prediction. Thus, we can conclude that the explanations provided by GNNExplainer and GANExplainer are correct while the explanation of Gem is incorrect. Furthermore, comparing the explanation of GNNExplainer and GANExplainer, we note the explanation of GANExplainer provides a more complete explanation.

\section{Conclusion}
In this paper, we propose GANExplainer, an explainer based on GAN, which can create accurate and real explanations for any graph neural network. Specifically, GANExplainer is model-agnostic, does not rely on the linear-independence assumption of the explained features, and does not require knowledge of the internal structure of the target GNN. Furthermore, GANExplainer can be applied in inductive settings to explain the whole GNN model. Our findings are consistent across diverse datasets and graph learning tasks.



\bibliographystyle{splncs04}

\bibliography{reference.bib}

\begin{thebibliography}{10}
\providecommand{\url}[1]{\texttt{#1}}
\providecommand{\urlprefix}{URL }
\providecommand{\doi}[1]{https://doi.org/#1}

\bibitem{BajajCXPWLZ21}
Bajaj, M., Chu, L., Xue, Z.Y., Pei, J., Wang, L., Lam, P.C., Zhang, Y.: Robust
  counterfactual explanations on graph neural networks. In: Ranzato, M.,
  Beygelzimer, A., Dauphin, Y.N., Liang, P., Vaughan, J.W. (eds.) Advances in
  Neural Information Processing Systems 34: Annual Conference on Neural
  Information Processing Systems 2021, NeurIPS 2021, December 6-14, 2021,
  virtual. pp. 5644--5655 (2021),
  \url{https://proceedings.neurips.cc/paper/2021/hash/2c8c3a57383c63caef6724343eb62257-Abstract.html}

\bibitem{chen2021causal}
Chen, D., Zhao, H., He, J., Pan, Q., Zhao, W.: An {Causal} {XAI} {Diagnostic}
  {Model} for {Breast} {Cancer} {Based} on {Mammography} {Reports}. In: 2021
  {IEEE} {International} {Conference} on {Bioinformatics} and {Biomedicine}
  ({BIBM}). pp. 3341--3349 (Dec 2021). \doi{10.1109/BIBM52615.2021.9669648}

\bibitem{EhsaniMF18}
Ehsani, K., Mottaghi, R., Farhadi, A.: Segan: Segmenting and generating the
  invisible. In: 2018 {IEEE} Conference on Computer Vision and Pattern
  Recognition, {CVPR} 2018, Salt Lake City, UT, USA, June 18-22, 2018. pp.
  6144--6153. Computer Vision Foundation / {IEEE} Computer Society (2018).
  \doi{10.1109/CVPR.2018.00643}

\bibitem{GAN_2014}
Goodfellow, I.J., Pouget{-}Abadie, J., Mirza, M., Xu, B., Warde{-}Farley, D.,
  Ozair, S., Courville, A.C., Bengio, Y.: Generative adversarial nets. In:
  Ghahramani, Z., Welling, M., Cortes, C., Lawrence, N.D., Weinberger, K.Q.
  (eds.) Advances in Neural Information Processing Systems 27: Annual
  Conference on Neural Information Processing Systems 2014, December 8-13 2014,
  Montreal, Quebec, Canada. pp. 2672--2680 (2014)

\bibitem{huang_graphlime:_2022}
Huang, Q., Yamada, M., Tian, Y., Singh, D., Chang, Y.: {GraphLIME}: {Local}
  {Interpretable} {Model} {Explanations} for {Graph} {Neural} {Networks}. IEEE
  Transactions on Knowledge and Data Engineering pp.~1--6 (2022).
  \doi{10.1109/TKDE.2022.3187455}

\bibitem{jiang2022graph}
Jiang, W., Luo, J.: Graph neural network for traffic forecasting: A survey.
  Expert Systems with Applications  \textbf{207},  117921 (Nov 2022).
  \doi{10.1016/j.eswa.2022.117921}

\bibitem{kazius2005derivation}
Kazius, J., McGuire, R., Bursi, R.: Derivation and validation of toxicophores
  for mutagenicity prediction. Journal of Medicinal Chemistry  \textbf{48}(1),
  312--320 (Jan 2005). \doi{10.1021/jm040835a}

\bibitem{kipf2016semi}
Kipf, T.N., Welling, M.: Semi-supervised classification with graph
  convolutional networks  (2017)

\bibitem{LiW16}
Li, C., Wand, M.: Precomputed real-time texture synthesis with markovian
  generative adversarial networks. In: Leibe, B., Matas, J., Sebe, N., Welling,
  M. (eds.) Computer Vision - {ECCV} 2016 - 14th European Conference,
  Amsterdam, The Netherlands, October 11-14, 2016, Proceedings, Part {III}.
  Lecture Notes in Computer Science, vol.~9907, pp. 702--716. Springer (2016).
  \doi{10.1007/978-3-319-46487-9\_43},
  \url{https://doi.org/10.1007/978-3-319-46487-9\_43}

\bibitem{GNNs_Explainers_Robustness}
Li, Y., Verma, S., Yang, S., Zhou, J., Chen, F.: Are graph neural network
  explainers robust to graph noises? In: Aziz, H., Corr{\^{e}}a, D., French, T.
  (eds.) {AI} 2022: Advances in Artificial Intelligence - 35th Australasian
  Joint Conference, {AI} 2022, Perth, WA, Australia, December 5-8, 2022,
  Proceedings. Lecture Notes in Computer Science, vol. 13728, pp. 161--174.
  Springer (2022). \doi{10.1007/978-3-031-22695-3\_12},
  \url{https://doi.org/10.1007/978-3-031-22695-3\_12}

\bibitem{yiqiaosurvey}
Li, Y., Zhou, J., Verma, S., Chen, F.: A survey of explainable graph neural
  networks: Taxonomy and evaluation metrics. CoRR  \textbf{abs/2207.12599}
  (2022). \doi{10.48550/arXiv.2207.12599},
  \url{https://doi.org/10.48550/arXiv.2207.12599}

\bibitem{lin2021generative}
Lin, W., Lan, H., Li, B.: Generative causal explanations for graph neural
  networks. In: Meila, M., Zhang, T. (eds.) Proceedings of the 38th
  International Conference on Machine Learning, {ICML} 2021, 18-24 July 2021,
  Virtual Event. Proceedings of Machine Learning Research, vol.~139, pp.
  6666--6679. {PMLR} (2021),
  \url{http://proceedings.mlr.press/v139/lin21d.html}

\bibitem{DBLP_LucicHTRS22}
Lucic, A., ter Hoeve, M.A., Tolomei, G., de~Rijke, M., Silvestri, F.:
  Cf-gnnexplainer: Counterfactual explanations for graph neural networks. In:
  Camps{-}Valls, G., Ruiz, F.J.R., Valera, I. (eds.) International Conference
  on Artificial Intelligence and Statistics, {AISTATS} 2022, 28-30 March 2022,
  Virtual Event. Proceedings of Machine Learning Research, vol.~151, pp.
  4499--4511. {PMLR} (2022),
  \url{https://proceedings.mlr.press/v151/lucic22a.html}

\bibitem{LuoCXYZC020}
Luo, D., Cheng, W., Xu, D., Yu, W., Zong, B., Chen, H., Zhang, X.:
  Parameterized explainer for graph neural network. In: Proceedings of the 34th
  {International} {Conference} on {Neural} {Information} {Processing}
  {Systems}. pp. 19620--19631. {NIPS}'20, Curran Associates Inc., Red Hook, NY,
  USA (Dec 2020)

\bibitem{QiaoZXT19}
Qiao, T., Zhang, J., Xu, D., Tao, D.: Mirrorgan: Learning text-to-image
  generation by redescription. In: {IEEE} Conference on Computer Vision and
  Pattern Recognition, {CVPR} 2019, Long Beach, CA, USA, June 16-20, 2019. pp.
  1505--1514. Computer Vision Foundation / {IEEE} (2019).
  \doi{10.1109/CVPR.2019.00160}

\bibitem{RadfordMC15}
Radford, A., Metz, L., Chintala, S.: Unsupervised representation learning with
  deep convolutional generative adversarial networks. In: Bengio, Y., LeCun, Y.
  (eds.) 4th International Conference on Learning Representations, {ICLR} 2016,
  San Juan, Puerto Rico, May 2-4, 2016, Conference Track Proceedings (2016),
  \url{http://arxiv.org/abs/1511.06434}

\bibitem{RGExplainer_ShanSZLL21}
Shan, C., Shen, Y., Zhang, Y., Li, X., Li, D.: Reinforcement learning enhanced
  explainer for graph neural networks. In: Ranzato, M., Beygelzimer, A.,
  Dauphin, Y.N., Liang, P., Vaughan, J.W. (eds.) Advances in Neural Information
  Processing Systems 34: Annual Conference on Neural Information Processing
  Systems 2021, NeurIPS 2021, December 6-14, 2021, virtual. pp. 22523--22533
  (2021)

\bibitem{VondrickPT16}
Vondrick, C., Pirsiavash, H., Torralba, A.: Generating videos with scene
  dynamics. In: Lee, D.D., Sugiyama, M., von Luxburg, U., Guyon, I., Garnett,
  R. (eds.) Advances in Neural Information Processing Systems 29: Annual
  Conference on Neural Information Processing Systems 2016, December 5-10,
  2016, Barcelona, Spain. pp. 613--621 (2016),
  \url{https://proceedings.neurips.cc/paper/2016/hash/04025959b191f8f9de3f924f0940515f-Abstract.html}

\bibitem{WaleWK08}
Wale, N., Watson, I.A., Karypis, G.: Comparison of descriptor spaces for
  chemical compound retrieval and classification. Knowl. Inf. Syst.
  \textbf{14}(3),  347--375 (2008). \doi{10.1007/s10115-007-0103-5}

\bibitem{wang2022hagen}
Wang, C., Lin, Z., Yang, X., Sun, J., Yue, M., Shahabi, C.: {HAGEN}:
  {Homophily}-{Aware} {Graph} {Convolutional} {Recurrent} {Network} for {Crime}
  {Forecasting}. vol.~36, pp. 4193--4200 (Jun 2022).
  \doi{10.1609/aaai.v36i4.20338}

\bibitem{RC_Explainer_Wang}
Wang, X., Wu, Y., Zhang, A., Feng, F., He, X., Chua, T.: Reinforced causal
  explainer for graph neural networks. CoRR  \textbf{abs/2204.11028} (2022).
  \doi{10.48550/arXiv.2204.11028},
  \url{https://doi.org/10.48550/arXiv.2204.11028}

\bibitem{multi_grained_WangWZHC21}
Wang, X., Wu, Y., Zhang, A., He, X., Chua, T.: Towards multi-grained
  explainability for graph neural networks. In: Ranzato, M., Beygelzimer, A.,
  Dauphin, Y.N., Liang, P., Vaughan, J.W. (eds.) Advances in Neural Information
  Processing Systems 34: Annual Conference on Neural Information Processing
  Systems 2021, NeurIPS 2021, December 6-14, 2021, virtual. pp. 18446--18458
  (2021),
  \url{https://proceedings.neurips.cc/paper/2021/hash/99bcfcd754a98ce89cb86f73acc04645-Abstract.html}

\bibitem{wu2020comprehensive}
Wu, Z., Pan, S., Chen, F., Long, G., Zhang, C., Yu, P.S.: A comprehensive
  survey on graph neural networks. IEEE Transactions on Neural Networks and
  Learning Systems  \textbf{32}(1),  4--24 (Jan 2021).
  \doi{10.1109/TNNLS.2020.2978386}

\bibitem{YangKBP17}
Yang, J., Kannan, A., Batra, D., Parikh, D.: {LR-GAN:} layered recursive
  generative adversarial networks for image generation. In: 5th International
  Conference on Learning Representations, {ICLR} 2017, Toulon, France, April
  24-26, 2017, Conference Track Proceedings. OpenReview.net (2017),
  \url{https://openreview.net/forum?id=HJ1kmv9xx}

\bibitem{abs-1903-03894}
Ying, R., Bourgeois, D., You, J., Zitnik, M., Leskovec, J.: {GNN} explainer:
  {A} tool for post-hoc explanation of graph neural networks. CoRR
  \textbf{abs/1903.03894} (2019)

\bibitem{yuan_xgnn:_2020}
Yuan, H., Tang, J., Hu, X., Ji, S.: {XGNN}: {Towards} {Model}-{Level}
  {Explanations} of {Graph} {Neural} {Networks}. In: Proceedings of the 26th
  {ACM} {SIGKDD} {International} {Conference} on {Knowledge} {Discovery} \&
  {Data} {Mining}. pp. 430--438. ACM, Virtual Event CA USA (Aug 2020).
  \doi{10.1145/3394486.3403085}

\bibitem{XGNN_YuanTHJ20}
Yuan, H., Tang, J., Hu, X., Ji, S.: {XGNN:} towards model-level explanations of
  graph neural networks. In: Gupta, R., Liu, Y., Tang, J., Prakash, B.A. (eds.)
  {KDD} '20: The 26th {ACM} {SIGKDD} Conference on Knowledge Discovery and Data
  Mining, Virtual Event, CA, USA, August 23-27, 2020. pp. 430--438. {ACM}
  (2020). \doi{10.1145/3394486.3403085},
  \url{https://doi.org/10.1145/3394486.3403085}

\bibitem{zhou2021evaluating}
Zhou, J., Gandomi, A.H., Chen, F., Holzinger, A.: Evaluating the quality of
  machine learning explanations: A survey on methods and metrics. Electronics
  \textbf{10}(5), ~593 (Jan 2021). \doi{10.3390/electronics10050593}

\end{thebibliography}

\end{document}